\documentclass[letterpaper]{article} 
\usepackage{aaai2026}  
\usepackage{times}  
\usepackage{helvet}  
\usepackage{courier}  
\usepackage[hyphens]{url}  
\usepackage{graphicx} 
\usepackage{natbib}  
\usepackage{caption} 
\usepackage{booktabs}
\usepackage{multirow}
\usepackage{siunitx}
\usepackage{relsize}
\usepackage{amssymb}
\usepackage{subcaption}
\usepackage{amsmath}
\usepackage{cleveref}

\title{Griffin: Aerial-Ground Cooperative Detection and Tracking \\ Dataset and Benchmark
}
\author{
    Jiahao Wang\textsuperscript{\rm 1},
    Xiangyu Cao\textsuperscript{\rm 1}, 
    Jiaru Zhong\textsuperscript{\rm 2}, 
    Yuner Zhang\textsuperscript{\rm 3}, 
    Zeyu Han\textsuperscript{\rm 1},
    Haibao Yu\textsuperscript{\rm 4},\\ 
    Chuang Zhang\textsuperscript{\rm 1},
    Lei He\textsuperscript{\rm 1},
    Shaobing Xu\textsuperscript{\rm 1}
    \thanks{Corresponding authors: Jianqiang Wang and Shaobing Xu (wjqlws@tsinghua.edu.cn, shaobxu@tsinghua.edu.cn)}
    \addtocounter{footnote}{-1},
    Jianqiang Wang\textsuperscript{\rm 1}
    \footnotemark
    \thanks{This work was supported 
    by the National Natural Science Foundation of China for the Science Fund for Creative Research Groups (No. 52221005) and the Key Project (No. 52131201).
    }
}
\affiliations{
    \textsuperscript{\rm 1}Tsinghua University.
    \textsuperscript{\rm 2}The Hong Kong Polytechnic University.\\
    \textsuperscript{\rm 3}University of Pennsylvania.
    \textsuperscript{\rm 4}The University of Hong Kong.

}

\begin{document}

\maketitle

\begin{abstract}
While cooperative perception can overcome the limitations of single-vehicle systems, the practical implementation of vehicle-to-vehicle and vehicle-to-infrastructure systems is often impeded by significant economic barriers.
Aerial-ground cooperation (AGC), which pairs ground vehicles with drones, presents a more economically viable and rapidly deployable alternative.
However, this emerging field has been held back by a critical lack of high-quality public datasets and benchmarks.
To bridge this gap, we present \textit{Griffin}, a comprehensive AGC 3D perception dataset, featuring over 250 dynamic scenes (37k+ frames).
It incorporates varied drone altitudes (20-60m), diverse weather conditions, realistic drone dynamics via CARLA-AirSim co-simulation, and critical occlusion-aware 3D annotations.
Accompanying the dataset is a unified benchmarking framework for cooperative detection and tracking, with protocols to evaluate communication efficiency, altitude adaptability, and robustness to communication latency, data loss and localization noise.
By experiments through different cooperative paradigms, we demonstrate the effectiveness and limitations of current methods and provide crucial insights for future research.
The dataset and codes are available at \url{https://github.com/wang-jh18-SVM/Griffin}.
\end{abstract}


\section{Introduction}

\label{sec:introduction}


While significant progress has been made in autonomous driving technologies, current single-vehicle systems still struggle with fundamental challenges of severe occlusions and limited field-of-view in complex environments.
To address these limitations, cooperative perception strategies, including vehicle-to-vehicle (V2V) and vehicle-to-infrastructure (V2I), have emerged, offering substantial improvements in perception capabilities.
Nevertheless, their practical implementation often requires significant infrastructure investments and widespread adoption of connected vehicles, which can present substantial economic barriers.
In contrast, vehicle-to-drone or so called, aerial-ground cooperative (AGC) systems, which integrate unmanned aerial vehicles (UAVs) with ground vehicles, provide a unique alternative.
They leverage on-demand drone deployment and an unobstructed bird's-eye view, offering a flexible, economical solution for critical environments, including smart cities, emergency response, and security patrols.

\begin{figure}[t]
    \centering
    \includegraphics[width=\columnwidth]{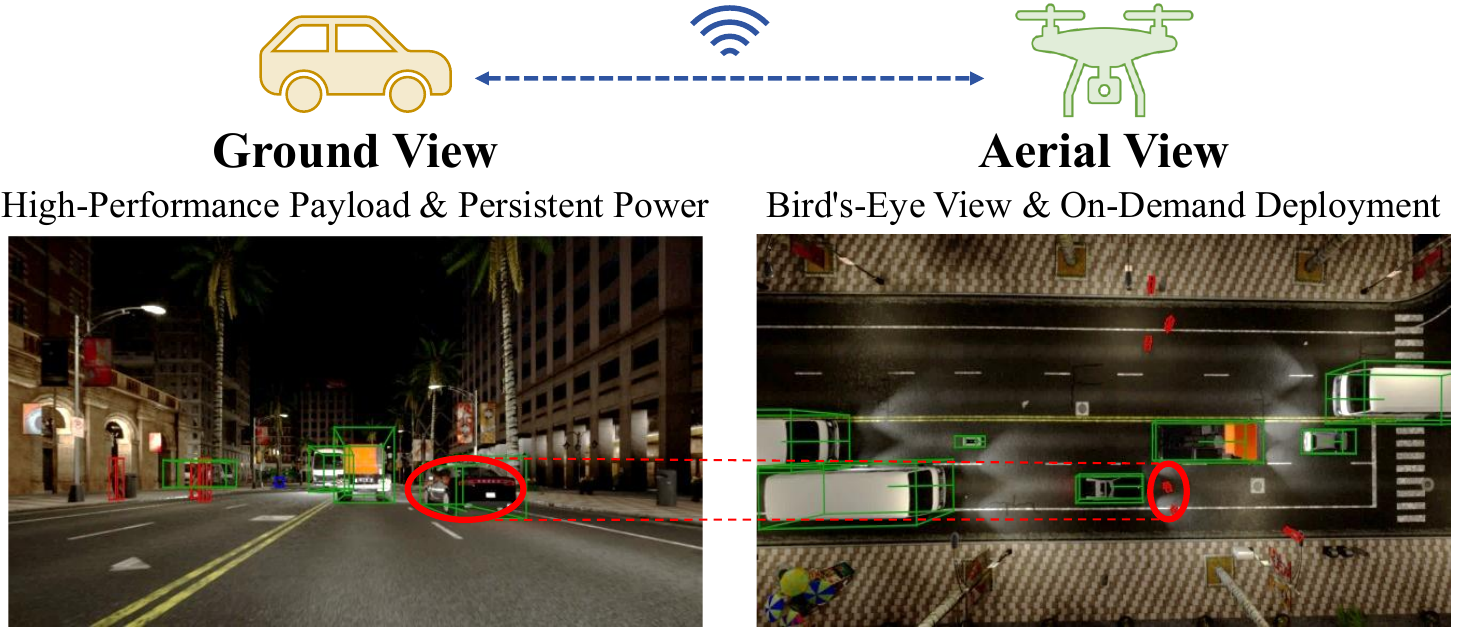}
    \caption{Motivation for aerial-ground cooperative perception.
    AGC provides a flexible alternative to fixed infrastructure by leveraging on-demand deployment and a unique bird's-eye view.
    In this example, the aerial view reveals pedestrians (red circle) occluded from the vehicle.
    }
    \label{fig:Motivation}
\end{figure}

\newcommand{\yes}{\checkmark}
\newcommand{\no}{$\times$}
\newcommand{\na}{--}
\newcommand{\citeConfYear}[1]{\small(#1)}

\begin{table*}[ht]
    \centering
    \setlength{\tabcolsep}{2.5pt} 
    \begin{tabular}{@{}cccccccccc@{}}
        \toprule
        \multirow{2}{*}{\textbf{Mode}} & \multirow{2}{*}{\textbf{Dataset}} & \multirow{2}{*}{\textbf{Source}} & \textbf{BBox} & \textbf{Cameras} & \textbf{Sequential} & \textbf{Occlusion} & \textbf{Realistic} & \textbf{Frames} & \textbf{Altitude} \\
        & & & \textbf{Type} & \textbf{per Agent} & \textbf{Tracking} & \textbf{Aware} & \textbf{Noise} & \textbf{(k)} & \textbf{(m)} \\
        \midrule
        \multirow{2}{*}{Veh-Veh} 
        & OPV2V\citeConfYear{ICRA 2022}        & Joint Sim & 3D & Multiple & \yes & \yes & \no  & 11    & \na \\
        & V2V4Real\citeConfYear{CVPR 2023}    & Real      & 3D & Multiple & \yes & \na  & \na  & 310   & \na \\
        \midrule
        \multirow{2}{*}{Veh-Inf} 
        & DAIR-V2X\citeConfYear{CVPR 2022}      & Real      & 3D & Single   & \no  & \na  & \na  & 22    & 20-25 \\
        & V2X-Seq\citeConfYear{CVPR 2023}     & Real      & 3D & Single   & \yes & \na  & \na  & 15    & 20-25 \\
        \midrule
        \multirow{4}{*}{Air-Air} 
        & \small{CoPerception-UAVs}\citeConfYear{NIPS 2022} & Joint Sim & 3D & Multiple & \yes & \no  & \yes  & 5.2   & 40,60,80 \\
        & UAV3D\citeConfYear{NIPS 2024}                & Joint Sim & 3D & Multiple & \yes & \no  & \no  & 20    & 60 \\
        & AeroCollab3D\citeConfYear{TGRS 2024}            & Joint Sim & 3D & Single   & \yes & \no  & \no  & 3.2   & 50 \\
        & Air-Co-Pred\citeConfYear{NIPS 2024}          & Sim       & 3D & Single   & \yes & \yes & \no  & 8.0   & 50 \\
        \midrule
        \multirow{5}{*}{Veh-Air} 
        & CoPeD\citeConfYear{RAL 2024}        & Real      & 2D & Single   & \no  & \na  & \na  & 203   & 2-10 \\
        & V2U-COO$^\dag$\citeConfYear{TGRS 2025} & Sim       & 3D & Single   & \no  & \no  & \yes  & 9.3   & 70, 80 \\
        & AGC-VUC$^\dag$\citeConfYear{Preprint}     & Real      & 3D & Multiple & \no  & \na  & \na  & 20    & 10-15 \\
        & AirV2X-Perception\citeConfYear{Preprint} & Joint Sim & 3D & Multiple & \yes & \no  & \yes  & 121.1 & 60-105 \\
        & \textbf{Griffin (Ours)} & \textbf{Joint Sim} & \textbf{3D} & \textbf{Multiple} & \textbf{\yes} & \textbf{\yes} & \textbf{\yes} & \textbf{37.7} & \textbf{20-60} \\
        \bottomrule
    \end{tabular}
    \caption{
        Comparison of representative cooperative perception datasets.
        Our dataset, \textit{Griffin}, is highlighted as the only one in the vehicle-aerial domain to support occlusion-aware annotations and realistic noise simulation.
        In the Source column, `Joint Sim' denotes co-simulation of CARLA and AirSim / SUMO; `Sim' uses  only CARLA.
        `Occlusion Aware' shows if annotation visibility is considered for simulated data.
        `Realistic Noise' indicates supports for simulating communication interference and localization errors.
        `Altitude' represents the height of infrastructure or drone sensors.
        $^\dag$Attributes are derived from the publications as the datasets are not released yet.
        Symbols: \yes\ (Supported), \no\ (Not Supported), \na\ (Not Applicable/Specified).
    }
    \label{tab:dataset-details}
\end{table*}

Despite the promising potential of AGC perception systems, relevant progress is hindered by a critical shortage of high-quality, representative benchmarks.
This gap stems from the inherent complexities of aerial-ground interaction, which pose significant challenges for both real-world data collection and high-fidelity simulation.
Unlike V2V or V2I systems, where sensors generally move on a horizontal plane, UAVs introduce complex motion with continuous changes in altitude, pitch, and roll, which disrupt precise cross-view correspondence and complicates the annotation and calibration process for real-world data.
Consequently, many efforts have turned to simulation to bypass these challenges.
However, existing simulation-based datasets still fail to replicate real-world complexities.
For instance, many datasets oversimplify the scene with ideal localization and communication~\cite{yeUAV3DLargescale3D2024,tianUCDNetMultiUAVCollaborative2024,wangDronesHelpDrones2024,wangAVCPNetAAVVehicleCollaborative2025,gaoAirV2XUnifiedAirGround2025},
or employ simplistic drone models with fixed orientation~\cite{wangDronesHelpDrones2024, wangAVCPNetAAVVehicleCollaborative2025} or constant altitudes~\cite{yeUAV3DLargescale3D2024, tianUCDNetMultiUAVCollaborative2024, wangDronesHelpDrones2024, wangAVCPNetAAVVehicleCollaborative2025}.
Furthermore, many of them~\cite{huWhere2commCommunicationefficientCollaborative2022,yeUAV3DLargescale3D2024,tianUCDNetMultiUAVCollaborative2024,wangAVCPNetAAVVehicleCollaborative2025,gaoAirV2XUnifiedAirGround2025} lack occlusion-aware annotations, resulting in labels for invisible objects.
Overcoming these shortcomings is crucial for developing robust AGC perception systems applicable to real-world environments.

Inspired by the Griffin, a mythical creature that unites the lion's terrestrial strength and the eagle's aerial dominance, we aim to harness the combined power of aerial and ground perspectives to overcome these challenges and enhance collaborative ability.
To this end, we present the following contributions for aerial-ground cooperative 3D perception:
\begin{itemize}
    \item \textbf{The Griffin Dataset}: We release \textit{Griffin}, an aerial-ground cooperative 3D perception dataset. It encompasses over 250 dynamic scenes (37K frames, 340K images) from CARLA-AirSim co-simulation, with instance-aware occlusion quantification, varying cruising altitudes, and realistic simulation of drone dynamics under various conditions. 
    \item \textbf{A Comprehensive Benchmark}: We present a benchmarking framework for evaluating aerial-ground cooperative 3D object detection and tracking. It includes implementations of classic baselines and provides a suite of metrics to evaluate accuracy, communication cost, and robustness under varying communication interference and localization errors. 
\end{itemize}

\section{Related Work}
\label{sec:related}

Cooperative perception research has been significantly propelled by datasets spanning various communication modes, as detailed in Table~\ref{tab:dataset-details}.
For V2V and V2I scenarios, benchmarks range from comprehensive simulations~\cite{xuOPV2VOpenBenchmark2022,liV2XsimMultiagentCollaborative2022} to real-world collections~\cite{yuDAIRV2XLargescaleDataset2022,xuV2V4RealRealworldLargescale2023,yuV2XseqLargescaleSequential2023,haoRCooperRealworldLargescale2024}.
In contrast, the cooperative perception datasets featuring aerial perspectives from UAVs remain limited. 
Existing works, such as UAV3D~\cite{yeUAV3DLargescale3D2024}, AeroCollab3D~\cite{tianUCDNetMultiUAVCollaborative2024}, and Air-Co-Pred~\cite{wangDronesHelpDrones2024} primarily focus on Air-Air cooperation and are often constrained by idealized communication and localization or simplified drone dynamics, such as fixed altitudes.

For the more challenging AGC scenarios, pioneering datasets have emerged, though with notable limitations.
CoPeD~\cite{zhouCoPeDAdvancingMultiRobotCollaborative2024}, while large-scale, targets low-altitude robot scenarios and employs unrefined, automatically generated annotations.
V2U-COO~\cite{wangUVCPNetUAVvehicleCollaborative2024,wangAVCPNetAAVVehicleCollaborative2025} relies on predefined UAV poses that lack realistic motion dynamics.
More recent efforts have also sought to advance the field but leave critical gaps.
AGC-Drive~\cite{houAGCDriveLargeScaleDataset2025} lacks tracking IDs for temporal analysis and is constrained to low-altitude flights,
while AirV2X~\cite{gaoAirV2XUnifiedAirGround2025} omits crucial occlusion-aware annotations.

To address these limitations, our work introduces \textit{Griffin}, a comprehensive solution designed to advance deployable aerial-ground perception systems.
Our dataset bridges the aforementioned gaps by providing realistic multi-agent dynamics through co-simulation, complete with occlusion-aware 3D annotations and tracking IDs across diverse altitude settings.
Furthermore, we introduce a robust benchmarking framework engineered to bridge the sim-to-real gap,
which allows a systematic robustness evaluation against controllable, real-world imperfections.

\begin{figure*}[t]
    \centering
    \includegraphics[width=0.9\linewidth]{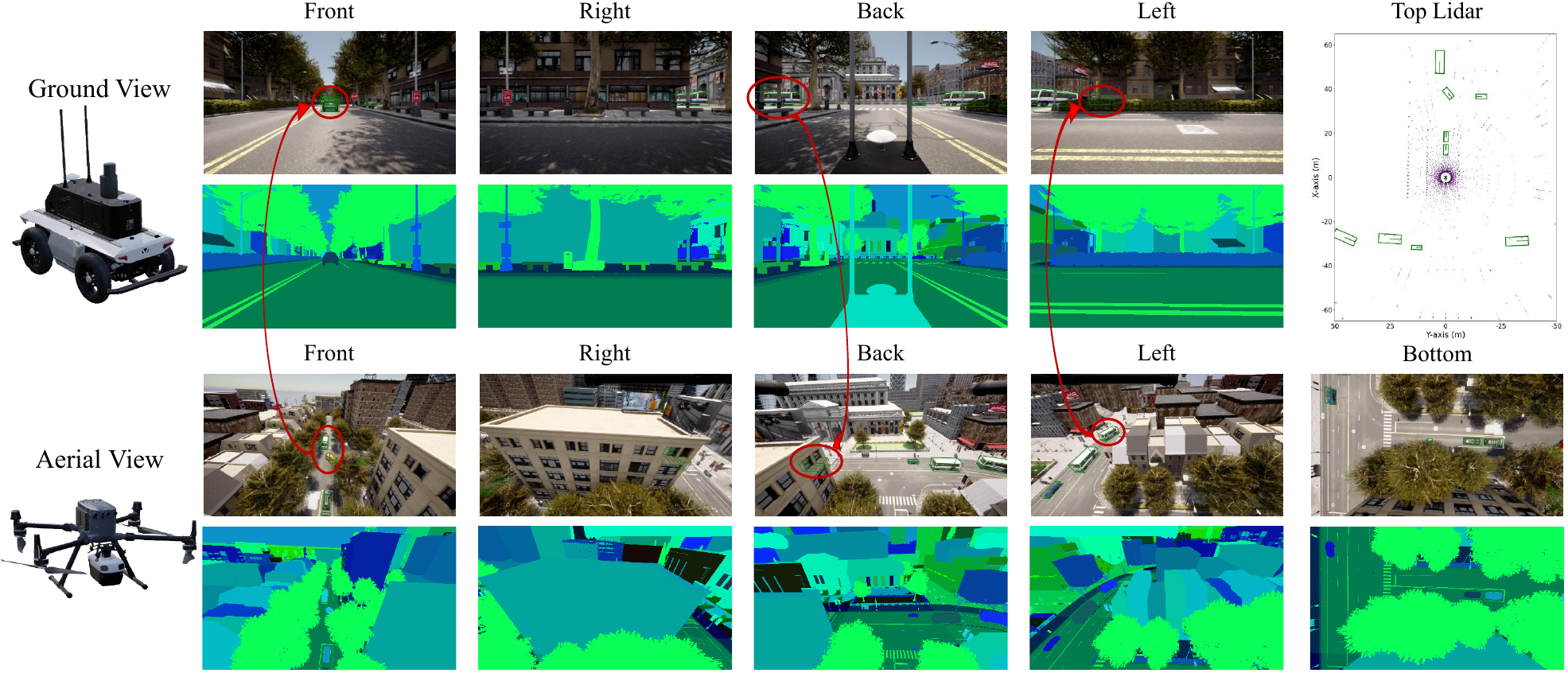}
    
    \caption{An example from \textit{Griffin} with visualized annotation.
    The ground vehicle platform is equipped with four cameras and one LiDAR, while the aerial drone platform has five cameras.
    We also provide instance segmentation ground truth, as shown in the lower row.
    Bounding boxes represent annotations from cooperative perspectives, indicating that one agent should be able to ‘see’ certain occluded objects after communication with the other.
    We use red circles and arrows to highlight those cases.
    }
    \label{fig:data_example}
\end{figure*}
\section{Data Setup}
\label{sec:data}

\subsection{Data Collection}
\begin{figure}[t]
    \centering
    \includegraphics[width=0.9\columnwidth]{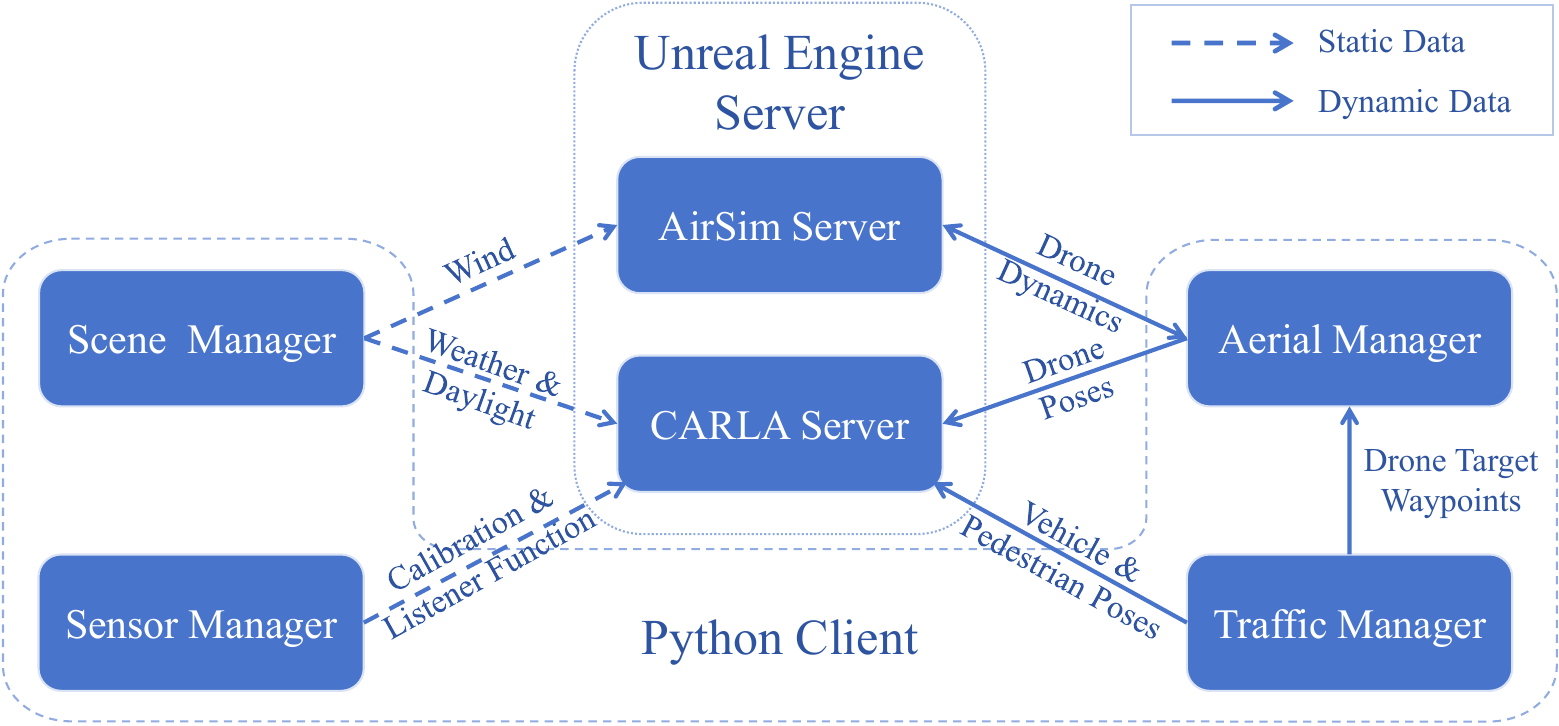} 
    \caption{Data collection framework.
    }
    \label{fig:data_collection_framework}
\end{figure}

To generate synchronized multi-agent scenarios, we employ a modular co-simulation framework built on CARLA~\cite{dosovitskiyCARLAOpenUrban2017} and AirSim~\cite{shahAirSimHighFidelityVisual2018}, as detailed in Figure~\ref{fig:data_collection_framework}.
This architecture leverages CARLA for its rich environmental maps, dynamic traffic flows, and high-fidelity sensors, complemented by AirSim's realistic, physics-based modeling of drone dynamics.

\noindent\textbf{Sensor Configuration.}
The sensor suites for both ground and aerial platforms are carefully designed to balance perceptual capabilities with platform-specific constraints.
As illustrated in Figure~\ref{fig:data_example}, the ground platform features a multi-modal sensor suite with four wide field-of-view (FoV) RGB cameras (108.8°, 1920×1080 resolution) and an 80-beam LiDAR operating at 10 Hz with a vertical FoV from -25° to 15°.
In contrast, strict size, weight, and power (SWaP) constraints for the aerial platform necessitate a vision-centric configuration without LiDAR. 
Consequently, the aerial platform employs five downward-oriented cameras with sensor specifications matching those of the ground vehicle.

\begin{figure}[t]
    \centering
        \includegraphics[width=0.7\columnwidth]{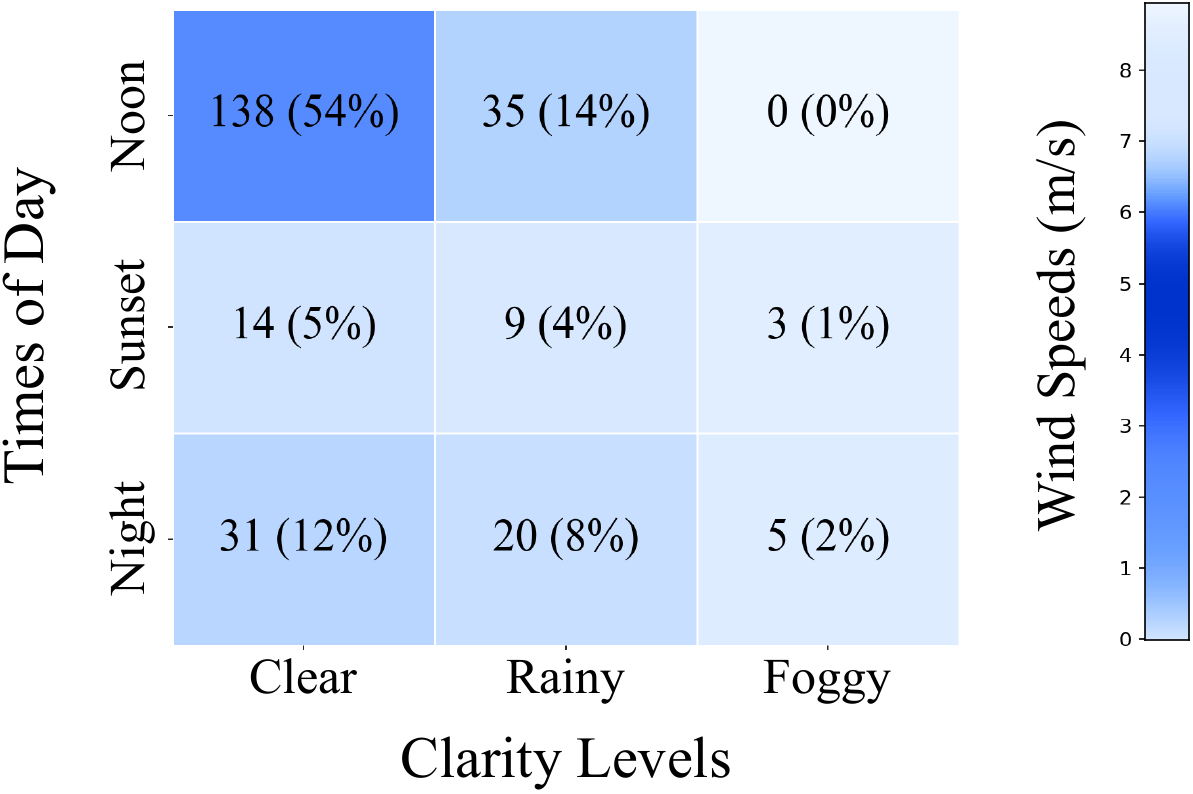}
    \caption{Weather distribution of scene clips.
    The dataset encompasses a variety of weather and lighting conditions.
    Following real-world patterns, certain combinations, such as fog at noon, are intentionally rare or absent.
    }
    \label{fig:Weather_Distribution}
\end{figure}

\noindent\textbf{Scene Diversity.}
To ensure robustness and generalizability, the dataset was collected across a wide range of simulated environments and conditions.
Data were captured from four representative CARLA maps—two urban (Town03, Town10HD) and two suburban (Town06, Town07)—with varying actor densities and vehicle speeds.
As detailed in Figure~\ref{fig:Weather_Distribution}, scenes feature diverse weather, including different times of day (noon, sunset, night), clarity levels (clear, rainy, foggy), and wind speeds (0–9 m/s).

Based on UAV cruising altitude, the dataset is organized into four categories.
The \textit{Griffin-Random} features the widest altitude range, from 20 to 60 meters above the vehicle.
In contrast, \textit{Griffin-25m}, \textit{Griffin-40m}, and \textit{Griffin-55m} focus on specific altitude bands at $25 \pm 2$m, $40 \pm 2$m, and $55 \pm 2$m, respectively.
In total, the dataset comprises 255 scene clips, each lasting approximately 15 seconds, totaling over 37.7k samples, 339.3k images, and 914.8k 3D annotations.

\begin{figure}[t]
    \centering
        \includegraphics[width=\columnwidth]{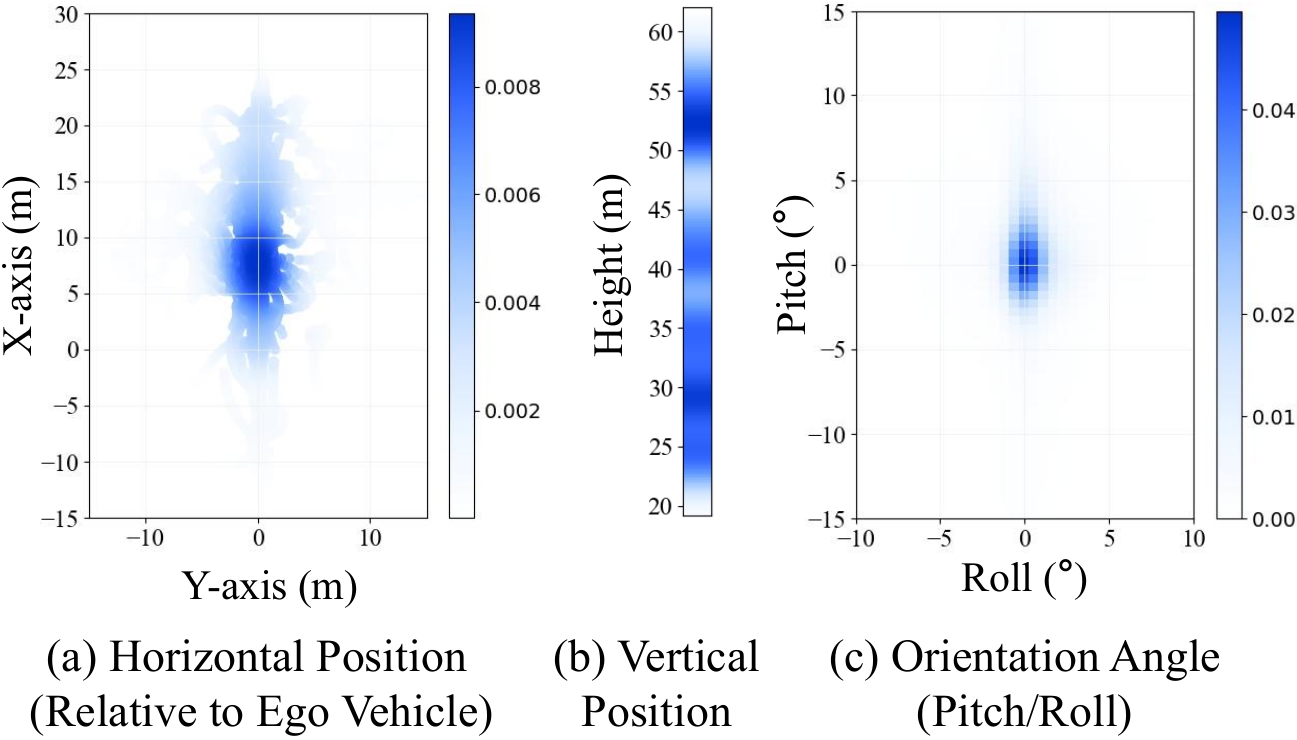}
    \caption{UAV pose distribution of \textit{Griffin-Random}.
    }
    \label{fig:UAV Pose Distribution Analysis}
\end{figure}

Furthermore, we designed multiple collaboration modes by varying the horizontal and vertical distances between the ground vehicle and UAV, creating diverse relative positioning patterns.
As shown for \textit{Griffin-Random} in Figure~\ref{fig:UAV Pose Distribution Analysis}, the drone is typically positioned several meters ahead of the vehicle, serving as a forward scout.
The realism of our simulation is further reflected in the drone's orientation angles.
The distributions for roll and pitch angles show variance around zero rather than being sharply peaked, which demonstrates the drone's continuous micro-adjustments to reach acceleration targets and maintain stability against simulated wind.

\subsection{Data Post-Processing}

\begin{table}[!b]
    \centering
    \footnotesize
    \begin{tabular}{c c c c}
        \toprule
        \textbf{Name} & \textbf{Category} & \textbf{Type} & \textbf{Origin} \\
        \midrule
        \multirow{1}{*}{World} & Geodetic & ENU (R) & Fixed reference point \\
        \midrule
        \multirow{2}{*}{Ego} & Drone & FLU (R) & Drone center \\
        & Vehicle & FLU(R) & Vehicle center \\
        \midrule
        \multirow{2}{*}{Sensor} & Camera & RDF (R) & Camera optical center \\
        & LiDAR & FLU (R) & LiDAR center \\
        \midrule
        \multirow{2}{*}{Simulator} & CARLA & ESU (L) & Fixed reference point \\
        & AirSim & NED (R) & Fixed reference point\\
        \bottomrule
    \end{tabular}
    \caption{Different coordinate systems.
    Axis direction: ENU (East-North-Up), FLU (Forward-Left-Up), RDF (Right-Down-Forward), RD (Right-Down), ESU (East-South-Up), NED (North-East-Down).
    Handedness: R (Right-handed), L (Left-handed)} 
    \label{table:Different_Coordinate_Systems}
\end{table}

\noindent\textbf{Spatio-Temporal Alignment.}
The \textit{Griffin} dataset involves four coordinate system categories: world, ego, sensor, and simulator, as detailed in Table~\ref{table:Different_Coordinate_Systems}.
Our spatial alignment pipeline converts all simulator-native 3D annotations into a unified right-handed coordinate system.
Dual output formats are available that support both the ego-centric KITTI benchmark~\cite{geigerVisionMeetsRobotics2013} and the global-reference NuScenes benchmark~\cite{caesarNuScenesMultimodalDataset2020}.

To ensure temporal consistency, CARLA's synchronous mode was activated during data recording, which guarantees that all data captured share a precise timestamp.

To bridge the sim-to-real gap, we also provide a code interface to inject user-specified real-world imperfections during training and inference, including localization errors, communication latency, and packet loss, enabling a full evaluation of model robustness.

\noindent\textbf{Occlusion-Aware Annotation.}
\textit{Griffin} provides high-quality, frame-by-frame 3D annotations for six object categories: pedestrian, car, truck, bus, motorcycle, and bicycle.
Each annotation includes a category label, a persistent ID, a calculated visibility rate, and a 9-DoF bounding box defined by x, y, z, length, width, height, roll, pitch, and yaw.

\begin{figure}[t]
    \centering
    \begin{subfigure}[b]{0.49\columnwidth}
        \centering
        \includegraphics[width=\linewidth]{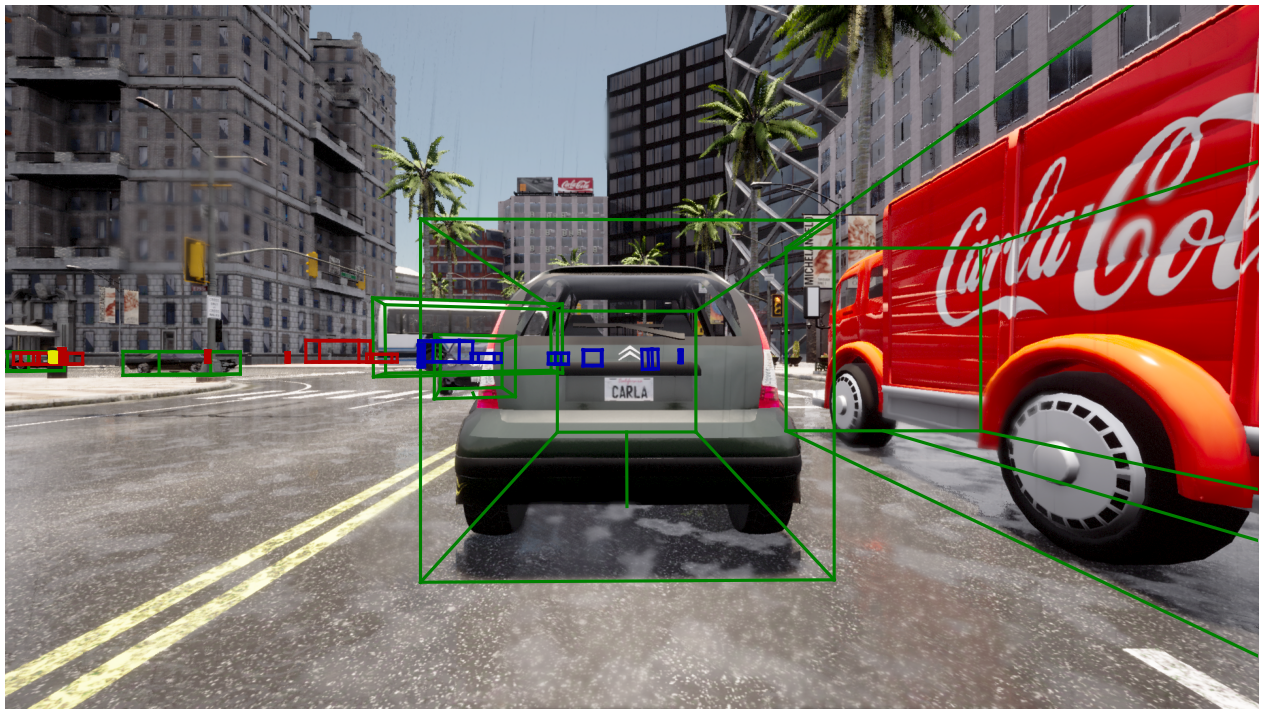} 
        \caption{Vehicle Front View}
        \label{fig:occ_aware_veh}
    \end{subfigure}
    \begin{subfigure}[b]{0.49\columnwidth}
        \centering
        \includegraphics[width=\linewidth]{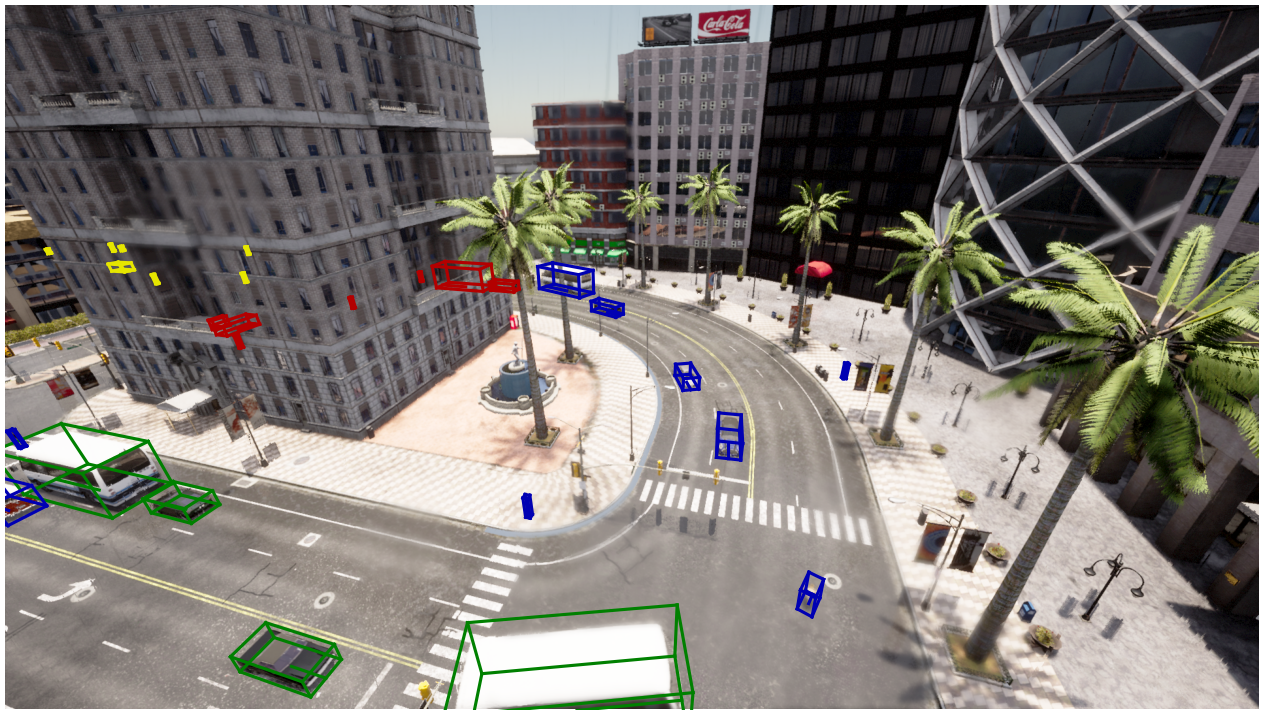} 
        \caption{Drone Front View}
        \label{fig:occ_aware_air}
    \end{subfigure}
    \caption{Unfiltered annotations regardless of visibility. 
    Boxes are color-coded by their visibility to show the necessity of our occlusion-aware filtering. 
    Green boxes represent targets visible to the vehicle, while blue boxes are those made visible by the drone's complementary view; both are retained as the ground truth for cooperative perception. 
    In contrast, many existing datasets only filter by distance (yellow boxes) and neglect heavily occluded targets (red boxes).
    }
    \label{fig:visibility}
\end{figure}

To address the critical challenge of judging occlusions, we developed a visibility quantification method that leverages CARLA's instance segmentation ground truth interface.
During data collection, RGB and segmentation images are recorded simultaneously using identical sensor configurations to ensure perfect spatio-temporal alignment.
In post-processing, we sample points within each ground-truth bounding box and project them onto the segmentation image.
Visibility rates for each agent are then calculated by comparing semantic categories and instance IDs of the sampled pixels against the corresponding target's.
Targets with low visibility from a single agent's perspective are filtered out to ensure annotation precision.
For the final cooperative perception ground truth, targets visible to either agent are retained, as illustrated with green and blue boxes in Figure~\ref{fig:visibility}.

\section{Tasks, Metrics, and Baselines}
\label{sec:task}

The \textit{Griffin} dataset is designed to support a variety of cooperative perception tasks, including but not limited to 3D object detection, tracking, motion prediction, and semantic segmentation.
In this paper, we narrow our focus to two fundamental visual tasks: 3D Object Detection and Tracking.

\subsection{AGC 3D Object Detection Task}

This task requires the detection of 3D objects within a predefined region of interest $R_g$ around the ego-vehicle using cooperative data from ground ($g$) and aerial ($a$) platforms.
For a perception timestamp $T_g$, inputs consist of image sequences $\{C_g(t_g) \mid t_g \leq T_g\}$ and $\{C_a(t_a) \mid t_a \leq T_a\}$, where $C(\cdot)$ denotes the capture function, along with the relative agent pose $M_{a_{T_a} \to g_{T_g}}$.
The desired output is a set of detections, each containing a 3D bounding box, a semantic label, and a confidence score.
The corresponding ground truth, $GT_{\text{detect}}$, is formulated by uniting the annotated objects visible to either agent and then filtering for those within the region of interest:
$$ GT_{\text{detect}} = \{ o \mid o \in GT_g \cup GT_a \text{ and } \text{center}(o) \in R_g \} $$
where each object $o$ is defined by its 3D bounding box and label, and $\text{center}(o)$ is the box's geometric center.

\subsection{AGC 3D Object Tracking Task}

The tracking task extends detection by requiring a model to maintain a unique and persistent identity for each object over time. Our framework supports two standard tracking paradigms: a \textit{joint tracking} approach that uses the same raw sensor inputs as the detection task, and a \textit{tracking-by-detection} approach that uses pre-computed detections as input.
In both cases, the output for each object must augment its 3D bounding box and label with a persistent tracking ID.
The corresponding ground truth, $GT_{\text{track}}$, is structured accordingly to include these IDs.

\subsection{Evaluation Metrics}

To ensure a comprehensive and standardized evaluation, we adopt established metrics from the NuScenes benchmark~\cite{caesarNuScenesMultimodalDataset2020} for 3D object detection and tracking, including Average Precision (AP) to assess detection quality and Average Multi-Object Tracking Accuracy (AMOTA) for tracking performance.
Beyond perception accuracy, we also assess the communication efficiency by quantifying the data transmitted in Bytes per second (BPS). 
This allows for a direct analysis of each method's trade-offs by comparing its performance improvement over a no-fusion baseline against its required communication bandwidth.

\subsection{Baseline Framework}
\label{sec:baseline}

\begin{figure}[t]
    \centering
    \includegraphics[width=1.0\columnwidth]{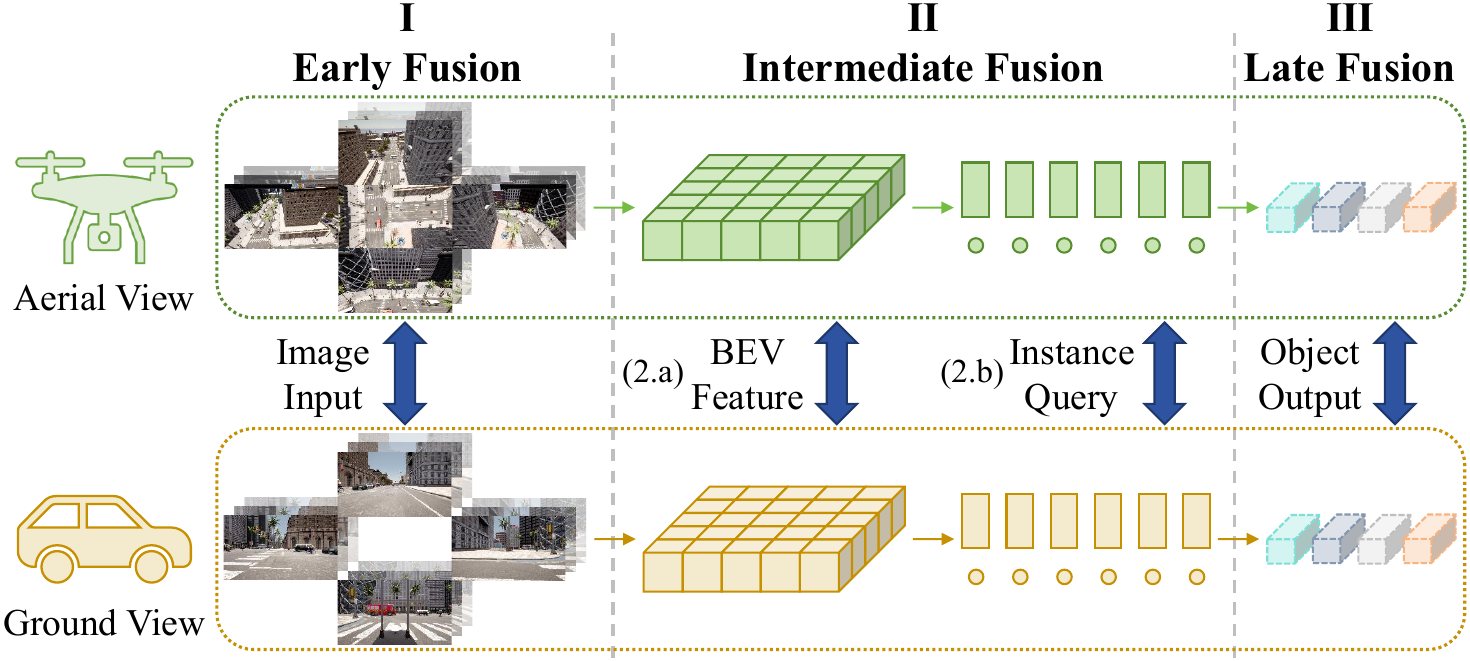}
    \caption{Overview of the cooperative fusion paradigms
    }
    \label{fig:fusion_stage}
\end{figure}

We implement and evaluate a series of baseline methods to establish performance references for AGC 3D object detection and tracking.
As shown in Figure~\ref{fig:fusion_stage}, existing cooperative perception methods can be categorized by fusion stages.
We provide implementations for all four major paradigms, all built upon a BEVFormer backbone~\cite{liBEVFormerLearningBirdseyeview2022}.

For intermediate fusion, which balances performance and communication costs, we evaluate four methods at two levels.
We implement V2X-ViT~\cite{xuV2XViTVehicletoeverythingCooperative2022} and Where2comm~\cite{huWhere2commCommunicationefficientCollaborative2022} as representative BEV-level fusion approaches, and UniV2X~\cite{yuEndtoEndAutonomousDriving2025} and CoopTrack~\cite{zhongCoopTrackExploringEndtoEnd2025} as instance-level ones.
These, along with standard Early and Late Fusion implementations, form a comprehensive framework to objectively compare different fusion strategies for AGC perception.

\section{Experiments}
\label{sec:experiments}

\begin{table*}[t]
    \centering
    \setlength{\tabcolsep}{3pt} 
    
    \newcommand{\gain}[1]{\small(#1)}
    \newcommand{\citefmt}[1]{\small(#1)}
    \newcommand{\best}[1]{\textbf{#1}}
    \newcommand{\costeffective}[1]{\underline{#1}}

    \begin{tabular}{c cc cc cc cc cc}
        \toprule
        \multirow{2}{*}[-0.5ex]{\textbf{Method}} & \multicolumn{2}{c}{\textbf{Griffin-25m}} & \multicolumn{2}{c}{\textbf{Griffin-40m}} & \multicolumn{2}{c}{\textbf{Griffin-55m}} & \multicolumn{2}{c}{\textbf{Griffin-Random}} & \multirow{2}{*}[-0.5ex]{\textbf{\begin{tabular}[c]{@{}c@{}}Comm. Cost \\ (BPS)\end{tabular}}} & \multirow{2}{*}[-0.5ex]{\textbf{\begin{tabular}[c]{@{}c@{}}Comp. Eff. \\ (FPS)\end{tabular}}} \\
        \cmidrule(lr){2-3} \cmidrule(lr){4-5} \cmidrule(lr){6-7} \cmidrule(lr){8-9}
        & {\textbf{AP}} & {\textbf{AMOTA}} & {\textbf{AP}} & {\textbf{AMOTA}} & {\textbf{AP}} & {\textbf{AMOTA}} & {\textbf{AP}} & {\textbf{AMOTA}} & & \\
        \midrule

        No Fusion & 0.375 & 0.365 & 0.341 & 0.363 & 0.335 & 0.359 & 0.459 & 0.481 & 0 & 8.10 \\
        \midrule
        
        \multirow{2}{*}{Early Fusion} & \best{0.607} & \best{0.670} & \best{0.503} & \best{0.555} & \best{0.483} & \best{0.522} & \best{0.583} & \best{0.649} & \multirow{2}{*}{$\num{3.11e8}$} & \multirow{2}{*}{5.17} \\
        & \gain{+0.232} & \gain{+0.305} & \gain{+0.162} & \gain{+0.192} & \gain{+0.148} & \gain{+0.163} & \gain{+0.124} & \gain{+0.168} & & \\
        \addlinespace

        V2X-ViT & 0.465 & 0.508 & 0.410 & 0.502 & 0.350 & 0.379 & 0.400 & 0.423 & \multirow{2}{*}{$\num{8.00e5}$} & \multirow{2}{*}{7.56} \\
        \citefmt{ECCV 2022} & \gain{+0.090} & \gain{+0.143} & \gain{+0.069} & \gain{+0.139} & \gain{+0.015} & \gain{+0.020} & \gain{-0.059} & \gain{-0.058} & & \\
        \addlinespace
        
        Where2Comm & 0.396 & 0.406 & 0.345 & 0.413 & 0.317 & 0.353 & 0.406 & 0.451 & \multirow{2}{*}{$\num{3.30e5}$} & \multirow{2}{*}{7.60} \\
        \citefmt{NIPS 2022} & \gain{+0.021} & \gain{+0.041} & \gain{+0.004} & \gain{+0.050} & \gain{-0.018} & \gain{-0.006} & \gain{-0.053} & \gain{-0.030} & & \\
        \addlinespace

        CoopTrack & 0.479 & 0.488 & \costeffective{0.396} & 0.446 & \costeffective{0.364} & \costeffective{0.402} & \costeffective{0.468} & \costeffective{0.490} & \multirow{2}{*}{$\num{1.17e5}$} & \multirow{2}{*}{6.23} \\
        \citefmt{ICCV 2025} & \gain{+0.104} & \gain{+0.123} & \gain{+0.055} & \gain{+0.083} & \gain{+0.029} & \gain{+0.043} & \gain{+0.009} & \gain{+0.009} & & \\
        \addlinespace

        UniV2X & 0.419 & 0.456 & 0.348 & 0.401 & 0.323 & 0.349 & 0.402 & 0.443 & \multirow{2}{*}{$\num{5.58e4}$} & \multirow{2}{*}{7.06} \\
        \citefmt{AAAI 2025} & \gain{+0.044} & \gain{+0.091} & \gain{+0.007} & \gain{+0.038} & \gain{-0.012} & \gain{-0.010} & \gain{-0.057} & \gain{-0.038} & & \\
        \addlinespace

        \multirow{2}{*}{Late Fusion} & \costeffective{0.378} & \costeffective{0.377} & 0.335 & \costeffective{0.391} & 0.306 & 0.332 & 0.375 & 0.400 & \multirow{2}{*}{$\num{1.56e3}$} & \multirow{2}{*}{6.83} \\
        & \gain{+0.003} & \gain{+0.012} & \gain{-0.006} & \gain{+0.028} & \gain{-0.029} & \gain{-0.027} & \gain{-0.084} & \gain{-0.081} & & \\
        \bottomrule
    \end{tabular}
    \caption{
        Model performance, communication cost, and computational efficiency.
        The Frames Per Second (FPS) values are measured on a single NVIDIA 3090 GPU.
        Parenthesized values show absolute gain over the No Fusion baseline.
        Bold values denote the best overall performance. 
        Underlined values indicate the highest gain-per-byte efficiency.
    }
    \label{tab:multi_datasets_performance}
\end{table*}

\begin{figure*}[t]
    \centering
    \includegraphics[width=1.0\linewidth]{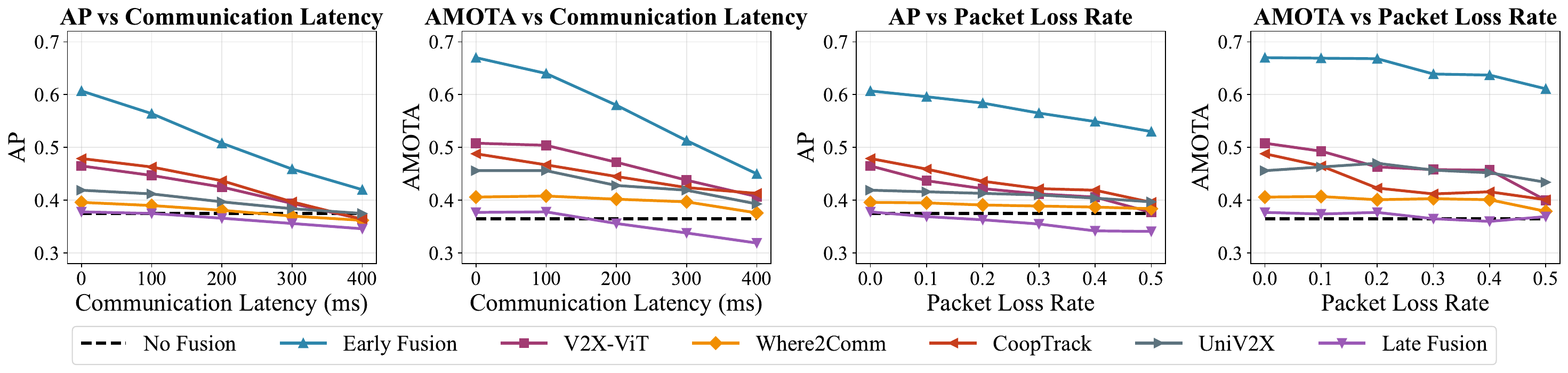}
    \caption{Robustness to communication interference.}
    \label{fig:communication_interference}
\end{figure*}

The detection and tracking performance of different fusion methods across the Griffin datasets are presented in Table~\ref{tab:multi_datasets_performance}, which details the AP and AMOTA scores alongside the associated communication costs.
The experiments reveal distinct performance characteristics for each fusion strategy.

\subsection{Performance Overview}
As expected, Early Fusion consistently establishes the upper performance bound, achieving the highest AP and AMOTA scores across all test conditions, albeit at an exceptionally high communication cost of $3.11 \times 10^8$ BPS (311 MB/s).
Conversely, Late Fusion represents the lower performance bound, offering only marginal gains and, in some cases, underperforming the No Fusion baseline.
However, due to its minimal communication cost of just $1.56 \times 10^3$ BPS, its modest improvements yield the highest gain-per-byte efficiency among all methods when a positive gain is achieved.

Intermediate fusion methods provide a balance between these extremes.
Among the methods with a communication cost under 1 MB/s, CoopTrack stands out, providing significant performance gains over the baseline at a moderate communication cost of $1.17\times10^{5}$ BPS.
V2X-ViT also demonstrates strong performance but requires a higher communication bandwidth due to its reliance on dense, scene-level BEV feature transmission.
Surprisingly, Where2comm and UniV2X yield unsatisfactory performance gains.
We attribute this to the inherent sparsity of targets from the drone's aerial perspective.
For methods like Where2comm, which use positive detections to generate spatial confidence maps for compressing BEV features, this sparsity can lead to the loss of valuable information.
Similarly, for sparse query-based methods like UniV2X, it can result in an insufficient number of matched positive samples during training, thereby limiting the model's learning capacity.
In contrast, although CoopTrack is also based on sparse object queries, its learnable instance association module leverages ground-truth matching relationships, enabling more effective cross-domain alignment and association.

\subsection{Generalization to Flight Altitude Changes}
The results in Table~\ref{tab:multi_datasets_performance} also demonstrate that the performance of cooperative perception methods is sensitive to UAV altitude, varying significantly across different flight heights.

\begin{figure*}[t]
    \centering
    \includegraphics[width=1.0\linewidth]{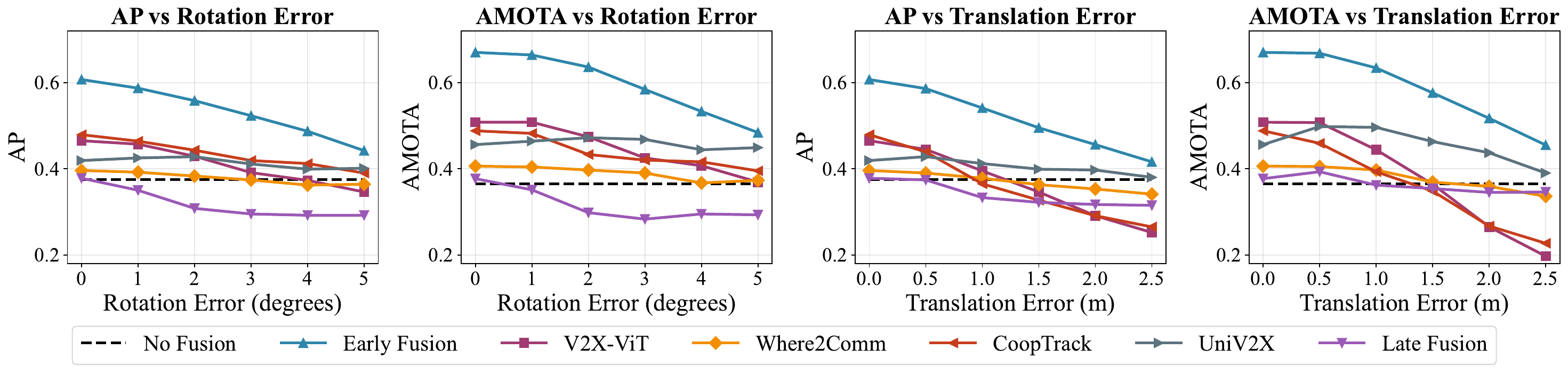}
    \caption{Robustness to localization error.}
    \label{fig:localization_error}
\end{figure*}

First, all cooperative methods achieve their highest performance gains on the low-altitude \textit{Griffin-25m} dataset.
As altitude increases in the \textit{Griffin-40m} and \textit{Griffin-55m} datasets, performance degrades across all methods.
This trend underscores a fundamental challenge in aerial perception: as altitude increases, the reduced apparent scale of ground targets makes them progressively more difficult to detect.
Interestingly, on the \textit{Griffin-Random} dataset, which features UAV altitudes ranging from 20m to 60m, most fusion methods perform worse than the No Fusion baseline.
This suggests that the varying altitudes, and the resulting inconsistencies in target scale and position, pose a critical challenge to the network's generalization capabilities, overriding the benefits from a lower average altitude or larger dataset size compared to \textit{Griffin-55m}.
These findings highlight the necessity for more adaptive fusion mechanisms that can accommodate dynamic UAV perspectives.

Furthermore, different fusion strategies exhibit distinct robustness profiles to altitude variations.
Instance-level methods prove more resilient, as CoopTrack is the only intermediate fusion method that maintains a performance advantage over the No Fusion baseline on the challenging \textit{Griffin-Random} dataset.
This gap can be attributed to their differing approaches to geometric transformation.
BEV-level fusion demands a rigid alignment of dense, geometrically-sensitive feature grids, a process vulnerable to the scale and perspective distortions caused by varying altitudes.
In contrast, instance-level methods flexibly transform the sparse 3D reference points associated with each object query while preserving their semantic features.
This decoupling of geometry and semantics makes the fusion process inherently more resilient to spatial inconsistencies.

\subsection{Communication Robustness}
To simulate real-world communication challenges, we evaluate the robustness of different fusion strategies against network imperfections, specifically communication latency and packet loss.
While existing studies~\cite{kutilaCV2X5GField2021} report typical latencies of 20-50 ms and packet loss rates (PLR) around 10\% under standard conditions, our evaluation intentionally explores a more demanding range—up to 400 ms latency and 50\% PLR.
This aggressive testing is designed to identify the failure points of each fusion strategy, which is particularly critical for aerial-ground scenarios where drones may encounter more severe signal interference and intermittent connectivity than their ground-based counterparts.

The results, illustrated in Figure~\ref{fig:communication_interference}, reveal inherent trade-offs between fusion paradigms and communication reliability.
Early Fusion, which transmits large volumes of raw image data, is highly susceptible to latency, with its AP score dropping by over 30\% at 400 ms.
In contrast, intermediate fusion methods exhibit better resilience.
While their detection performance surpasses the No Fusion baseline at latencies up to 200 ms, their tracking capabilities are even more robust, maintaining a consistent advantage across all tested latency levels up to 400 ms.

As for packet loss, its influence appears less detrimental than that of latency.
Even under severe conditions, such as a 50\% packet loss rate, most fusion methods maintain a performance advantage over the No Fusion baseline.
We hypothesize that this is because dropped packets lead to a loss of information, reducing potential performance gains, but do not introduce erroneous data that could generate additional false-positive signals.

\subsection{Localization Robustness}
We investigate the impact of localization errors by introducing noise into the UAV's transformation matrix, separately analyzing translation and rotation errors.
While existing benchmarks~\cite{xuV2XViTVehicletoeverythingCooperative2022,huWhere2commCommunicationefficientCollaborative2022,wangAVCPNetAAVVehicleCollaborative2025} often assess noise within narrow ranges (e.g., 0.6m or 1.0$^{\circ}$), we adopt a more challenging evaluation framework,
injecting Gaussian noise with standard deviations (std) of up to 2.5 meters for translation and 5 degrees for rotation.
This rigorous testing is crucial for understanding model reliability with severe GPS inaccuracies or calibration drifts.

As shown in Figure~\ref{fig:localization_error}, the performance of most cooperative methods is highly sensitive to both translation and rotation errors.
Among the intermediate fusion methods, V2X-ViT and CoopTrack suffer the most significant degradation, with their performance dropping below the No Fusion baseline when the translation error exceeds 1.5m. 
In contrast, UniV2X maintains a clear advantage over the No Fusion baseline across all tested error levels, and Where2comm also exhibits a more graceful degradation.
We attribute their superior robustness to their selective use of transmitted data, employing instance-level filtering or spatial confidence maps to down-weight unreliable signals. 

\section{Discussion and Conclusion}
\label{sec:conclusion}

This paper introduces the Griffin framework, a novel dataset and benchmark designed to accelerate research in aerial-ground cooperative 3D perception.
The experiments demonstrate the significant potential of this paradigm.
In favorable conditions, cooperative methods achieve substantial performance gains over single-agent baselines by resolving occlusions and expanding the effective field-of-view.

However, this work also underscores that the full potential of AGC is yet to be realized, as the performance gains of current fusion methods are fragile and highly dependent on idealized conditions.
Two primary challenges are identified: a strong sensitivity to the drone's flight altitude and a significant vulnerability to real-world imperfections like communication interference and localization errors.
Further analysis provides critical insights into these issues.
Instance-level fusion strategies, which exchange sparse object-centric information, seem more resilient to the perspective shifts from varying altitudes than their dense, BEV-level counterparts.
Meanwhile, the findings suggest that resilience to localization errors is directly linked to adaptive data filtering.
Methods that selectively fuse information—either through instance-level filtering or scene-level spatial confidence maps—proved more robust, highlighting that successful cooperation requires not just sharing data, but discerning which data to trust and fuse.
These findings indicate clear directions for future research.
Efforts should focus on developing altitude-adaptive and scale-aware fusion mechanisms capable of handling dynamic aerial viewpoints.
Advancing sparse fusion methods and creating dynamic trust mechanisms to weigh or filter erroneous signals will be critical.
Addressing these challenges is essential for developing robust aerial-ground cooperative systems that can be reliably deployed in unpredictable, real-world conditions.

Furthermore, our benchmark could be extended in several key areas.
We encourage the exploration of more advanced late-fusion strategies~\cite{chiuProbabilistic3DMultiObject2024}, which may boost performance at a minimal communication cost.
We also see value in benchmarking perception models specifically designed for the aerial domain,
which may provide more robust features for fusion.
Finally, future robustness analyses should investigate the impact of diverse weather conditions, evaluate performance under fair bandwidth-constrained scenarios, and incorporate asynchrony-robust methods.

\bibliography{THICV}

\clearpage
\appendix

This appendix provides supplementary details regarding our data collection framework, experimental setup, and additional ablation studies for key design choices.
Representative data samples are visualized in the supplementary video.

\section{Details about Data Collection}

\subsection{Data Collection Framework}

Our data collection framework is founded upon a modular architecture designed for generating synchronized, multi-agent scenarios.
The core of this framework is a server based on Unreal Engine 4 (UE4), which integrates the CARLA and AirSim  simulators.
This co-simulation setup is controlled by a Python client that features four specialized managers.
The Traffic Manager oversees vehicle path planning and control, while the Aerial Manager generates dynamic drone trajectories to serve as a forward scout.
Concurrently, the Scene Manager randomizes environmental parameters such as weather, time of day, and wind conditions to ensure scene diversity.
Finally, the Sensor Manager directs the acquisition and processing of multi-modal sensor data from both the ground and aerial platforms.

Once the simulation environment is initialized, the system loads predefined maps and populates them with dynamic traffic scenarios.
As vehicles navigate their routes autonomously, the framework captures all sensor data in a synchronous manner, which is then organized and stored for subsequent analysis.
\subsection{Sensor Configuration}

The perception systems for both the ground vehicle and aerial drone are modeled after real-world hardware to promote future sim-to-real transfer.
The components were carefully selected to balance perceptual capabilities with platform-specific operational constraints.

For the aerial platform, in particular, adherence to strict size, weight, and power (SWaP) constraints necessitates a vision-centric sensor suite, leading to the deliberate exclusion of LiDAR payloads.
Although certain existing datasets~\cite{houAGCDriveLargeScaleDataset2025} incorporate airborne LiDAR, its substantial weight and cost render it impractical for many real-world applications.
For instance, small civilian drones compatible with vehicle-integrated solutions typically have payload capacities under 1 kg, such as the system jointly developed by BYD Auto and DJI.
Even an industrial-grade drone like the DJI M350, with a higher payload capacity of 2.7 kg, remains constrained.
In our practical deployment, this capacity is fully allocated to essential components such as camera modules, an NVIDIA Orin carrier board for onboard processing, and communication equipment.
Consequently, our simulated aerial platform employs five downward-oriented cameras with specifications matching those on the ground vehicle, ensuring a realistic and deployable configuration.

\section{Experimental Setup}

This section details the experimental framework used to evaluate cooperative perception methods on the Griffin dataset.
We first provide a concise overview of related cooperative perception strategies to contextualize our choice of baselines.
We then describe the specific implementation of each baseline model, followed by the detailed training and evaluation parameters.

\subsection{Related Cooperative Perception Methods}

Cooperative perception methods are broadly categorized into three fusion strategies~\cite{caillotSurveyCooperativePerception2022, hanCollaborativePerceptionAutonomous2023,gaoVehicleRoadCloudCollaborativePerception2024} based on the stage at which data is shared: early, intermediate, and late fusion.
Early fusion~\cite{chenCooperCooperativePerception2019,arnoldCooperativePerception3D2022} integrates raw sensor data from multiple agents before feature extraction.
While this approach preserves the richest information, it imposes a prohibitive communication bandwidth overhead, making it impractical for most real-world applications but useful as a performance upper bound.
Late fusion~\cite{yuDAIRV2XLargescaleDataset2022,arnoldCooperativePerception3D2022} exchanges final perception outputs, such as 3D bounding boxes.
This method is extremely bandwidth-efficient but often yields limited performance gains under occlusion or ambiguous observations.
Intermediate fusion offers a balance by sharing partially processed network features and has become the focus of recent research.
One popular approach~\cite{huCollaborationHelpsCamera2023, liLearningDistilledCollaboration2021, wangV2VNetVehicletovehicleCommunication2020} involves fusing scene-level Bird's Eye View (BEV) features extracted from images or point clouds,
A more recent trend explores the transmission of sparse, instance-level features or queries~\cite{chenTransIFFInstanceLevelFeature2023, fanQUESTQueryStream2024}, which further reduces communication costs while maintaining robust performance.

Despite these advancements, methods explicitly designed for Aerial-Ground Cooperation (AGC) remain nascent, largely due to the scarcity of suitable public datasets.
Early explorations~\cite{minaeianVisionbasedTargetDetection2016,shenAeroNetEfficientRelative2023} include vision-based target localization and relative positioning using cooperative markers.
More recently, AVCPNet~\cite{wangAVCPNetAAVVehicleCollaborative2025} advances the field by fusing BEV features from both aerial and ground agents.
However, as noted in our main text, its validation on a dataset with simplified dynamics underscores the need for more realistic benchmarks like Griffin to drive further progress.

\subsection{Baseline Methods}


Following the established paradigm of V2I collaboration~\cite{zhongLeveragingTemporalContexts2024,yuEndtoEndAutonomousDriving2025}, our baseline framework is designed around structurally similar perception networks for both the UAV and the ground vehicle.
Different fusion modules were then integrated to assess the performance of each major cooperative strategy discussed in the main paper.

\noindent\textbf{Single-Agent Backbone:}
The foundation for all baselines is a strong single-agent perception model that functions independently on each platform.
Inspired by recent end-to-end driving models~\cite{huPlanningorientedAutonomousDriving2023,yuEndtoEndAutonomousDriving2025}, we employ BEVFormer~\cite{liBEVFormerLearningBirdseyeview2022} as the backbone to generate BEV features from multi-camera inputs.
Following the design of TrackFormer~\cite{meinhardtTrackFormerMultiobjectTracking2022}, instance queries are extracted from these features and propagated across frames to enable simultaneous object detection and tracking.
This model also serves as the "No Fusion" baseline in our experiments.

\noindent\textbf{Early Fusion:}
This method aggregates raw images from both the drone and vehicle.
After transforming drone camera extrinsics to the vehicle's coordinate system, all images are concatenated and processed by a unified BEVFormer backbone to produce a single, fused perception output.

\noindent\textbf{Late Fusion:}
This baseline operates on the final detection outputs.
Bounding boxes from the UAV are transmitted to the vehicle and fused using the Hungarian algorithm based on Euclidean distances.
Tracking is then performed on the fused detections using a standard tracking-by-detection approach AB3DMOT~\cite{weng3DMultiobjectTracking2020}, employing a Kalman filter for motion prediction and the Hungarian algorithm for cross-frame association.

\begin{figure}[t]
    \centering
    \includegraphics[width=1.0\columnwidth]{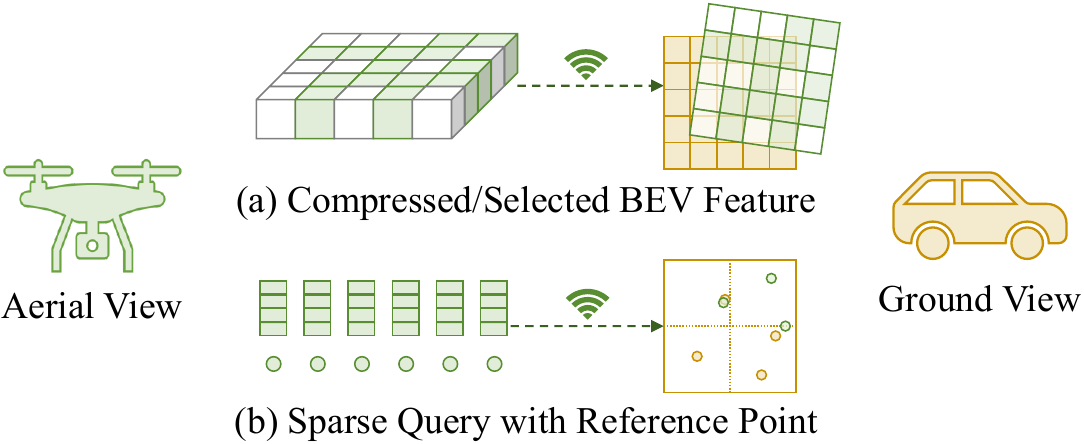}
    \caption{Two levels of intermediate fusion methods.}
    \label{fig:app_intermediate_fusion}
\end{figure}

\noindent\textbf{Intermediate Fusion (BEV-Level):}
We implement two variants to manage the high communication overhead caused by dense BEV features, as shown in Figure~\ref{fig:app_intermediate_fusion}.a.
The first, inspired by V2X-ViT~\cite{xuV2XViTVehicletoeverythingCooperative2022}, employs a ratio compression module (32:1 in our implementation) on the UAV's BEV features before transmission.
Upon reception, the vehicle decompresses the features, spatially aligns them with its own BEV map, and fuses them with a convolutional network.
The second approach, based on Where2comm~\cite{huWhere2commCommunicationefficientCollaborative2022}, employs an additional data reduction strategy.
It generates a spatial confidence map from object queries, which acts as a mask to transmit only the most salient feature grid cells.
Following the original method, this map is smoothed by a Gaussian filter (kernel size 5, sigma 1.0) and filtered by a confidence threshold of 0.001, which works in conjunction with a 4:1 dimensional compression.

\noindent\textbf{Intermediate Fusion (Instance-Level):}
We also implement two baselines that exchange sparse, object-centric queries, as illustrated in Figure~\ref{fig:app_intermediate_fusion}.b.
For UniV2X~\cite{yuEndtoEndAutonomousDriving2025}, each query comprises a feature vector and a 3D reference point.
The vehicle aligns these incoming queries using both explicit reference point projection and an implicit MLP-based method.
It then matches them with its own queries, fuses the matched pairs, and crucially, preserves high-confidence unmatched aerial queries to compensate for its own perceptual blind spots.
CoopTrack~\cite{zhongCoopTrackExploringEndtoEnd2025} enhances this process by separating query features into semantic and motion components, improving the alignment module with a latent transformation from STAR-Track~\cite{dollSTARTrackLatentMotion2024}, and utilizing a learnable module for more effective cross-agent query association.

\subsection{Implementation Details}

Experiments are conducted on the four splits of our dataset: \textit{Griffin-25m}, \textit{Griffin-40m}, \textit{Griffin-55m}, and \textit{Griffin-Random}, each divided into training and validation sets at an 8:2 ratio.
Following the conventions established by prior works such as Dair-V2X~\cite{yuDAIRV2XLargescaleDataset2022} and V2X-Seq~\cite{yuV2XseqLargescaleSequential2023}, we merge object categories into three classes for evaluation (cars, pedestrians, two-wheelers) and report primary results on the car class.
The perception range is set to a $102.4\text{m} \times 102.4\text{m}$ area centered on the ego-vehicle.
To ensure a fair comparison, all baseline methods are built on the same ResNet-50~\cite{heDeepResidualLearning2016} based BEVFormer backbone.
Input images are resized from $1920 \times 1080$ to $960 \times 540$.
All models are trained using the AdamW optimizer with a learning rate of $2 \times 10^{-4}$ and a batch size of 8, distributed across four NVIDIA 3090 GPUs.
Additional details regarding the software environment requirements are available in the code repository.

\section{More Experiments}

\begin{table}[!b]
    \centering
    \begin{tabular}{llcc}
        \toprule
        \textbf{Model} & \textbf{Annotation Filtering} & \textbf{AP $\uparrow$} & \textbf{AMOTA $\uparrow$} \\
        \midrule
        \multirow{2}{*}{Early Fusion} & \textbf{With (Baseline)} & \textbf{0.607} & \textbf{0.670} \\
        & Without & 0.586 & 0.636 \\
        \midrule
        \multirow{2}{*}{Vehicle Side} & \textbf{With (Baseline)} & \textbf{0.477} & \textbf{0.457} \\
        & Without & 0.412 & 0.433 \\
        \midrule
        \multirow{2}{*}{Drone Side} & \textbf{With (Baseline)} & \textbf{0.336} & \textbf{0.372} \\
        & Without & 0.326 & 0.356 \\
        \bottomrule
    \end{tabular}
    \caption{Impact of Occlusion-Aware Annotation on Griffin-25m. Training with unfiltered labels that include invisible objects degrades the performance of both cooperative and single-agent models.}
    \label{tab:occlusion_impact}
\end{table}

\subsection{Impact of Occlusion-Aware Labels}
This study validates the critical importance of our occlusion-aware annotation pipeline.
We conduct a comparative analysis by training models on two different sets of labels:
(1) the final, filtered occlusion-aware labels, 
and (2) unfiltered labels that include objects not visible to either agent.
As detailed in Table~\ref{tab:occlusion_impact}, all models were evaluated against the filtered ground truth.

The results unequivocally demonstrate that training with noisy, unfiltered labels leads to a tangible degradation in model accuracy.
For the Early Fusion model, performance drops from 0.607 to 0.586 in AP and from 0.670 to 0.636 in AMOTA.
We further extend this analysis to single-agent models (vehicle-only and drone-only) to show that unfiltered labels have a negative impact on individual perception as well.
The effect is particularly pronounced for the ground vehicle, which is more susceptible to occlusions, exhibiting a significant performance drop of 0.065 in AP and 0.024 in AMOTA.
These findings highlight that the precision of annotations is important and underscore the value of our occlusion-aware filtering process for developing robust perception systems.

\subsection{Impact of Drone's Surround-View Cameras}
This section presents an ablation study to demonstrate the value of the drone's panoramic camera setup.
We compare the performance of the drone-only perception model on Griffin-25m under two conditions: (1) using all five of the drone's surround-view cameras, and (2) using only the single bottom-facing camera.

As shown in Table~\ref{tab:camera_impact}, the results quantify the substantial performance improvement gained from the 360-degree aerial field-of-view.
When restricted to a single bottom-facing camera, the drone's performance declines, with AP dropping by 0.144 and AMOTA by 0.151.
This dramatic decrease underscores the expanded perceptual coverage provided by the multi-camera array, thereby justifying this key design feature of the Griffin dataset.

\begin{table}[t]
    \centering
    \begin{tabular}{lcc}
        \toprule
        \textbf{Drone Camera Setup} & \textbf{AP $\uparrow$} & \textbf{AMOTA $\uparrow$} \\
        \midrule
        \textbf{Surround-View (5 cameras)} & \textbf{0.336} & \textbf{0.372} \\
        Bottom-Only (1 camera) & 0.192 & 0.221 \\
        \bottomrule
    \end{tabular}
    \caption{Ablation study on the drone's camera configuration on Griffin-25m. The surround-view setup provides a significant performance advantage over a single camera.}
    \label{tab:camera_impact}
\end{table}


\end{document}